\documentclass[journal]{IEEEtran}
\usepackage{cite}
\usepackage{amsmath,amssymb,amsfonts}
\usepackage{algorithmic}
\usepackage{graphicx}
\usepackage{multirow}
\usepackage{makecell}
\usepackage{algorithm,algorithmic}
\usepackage{hyperref}
\usepackage{titlesec}
\usepackage{xcolor}
\hypersetup{hidelinks=true}
\usepackage{textcomp}
\usepackage{chngcntr}

\def\BibTeX{{\rm B\kern-.05em{\sc i\kern-.025em b}\kern-.08em
    T\kern-.1667em\lower.7ex\hbox{E}\kern-.125emX}}
\titlespacing*{\subsection}{0pt}{2pt}{2pt}

\begin{document}
%\raggedbottom
\title{Honest and Reliable Evaluation and Expert Equivalence Testing of Automated Neonatal Seizure Detection}
\author{Jovana Kljajic, John M. O'Toole, Robert Hogan, Tamara Skoric
\thanks{This work was supported by the EU-funded projects CA20124 - “Maximising the impact of multidisciplinary research in early diagnosis of neonatal brain injury" and CA24148 - “EEG101: Fundamentals of Open \& Rigorous EEG Science".}
\thanks{J. Kljajic and T. Skoric are with the Faculty of Technical Sciences, University of Novi Sad, 21000 Novi Sad, Serbia (e-mail: jovanakljajic18@uns.ac.rs, tamara.ceranic@gmail.com). }
\thanks{R. Hogan and J. M. O'Toole are with CergenX Ltd, Dublin, Ireland (e-mail: rhogan@cergenx.com, jotoole@cergenx.com).}}
%\thanks{Published version: \url{https://doi.org/10.1088/1873-4030/aea852}}

\maketitle

\begin{center}
    \small
    Published version:
    \url{https://doi.org/10.1088/1873-4030/aea852}
\end{center}

\begin{abstract}
Reliable evaluation of machine learning models for neonatal EEG seizure detection is a prerequisite for accurate performance assessment and comparison of methods. Model evaluation, however, is complicated by severe class imbalance and uncertain ground truth.  To address these limitations, we propose evidence-based reporting recommendations tailored to the specific challenges of neonatal seizure detection. Using real and synthetic seizure annotations, we assessed standard performance metrics, consensus strategies, and human–expert level equivalence tests under varying class imbalance, inter-rater agreement, and number of raters. Matthews and Pearson’s correlation coefficients outperformed the area under the receiver operating characteristic curve in reflecting performance under class imbalance. Consensus types are sensitive to the level of agreement and number of raters. Among the evaluated human--expert level equivalence tests, the multi-rater statistical Turing test using Fleiss' $\kappa$ demonstrated the most consistent performance and robustness under the tested experimental conditions. Based on our findings, we recommend reporting: (1) at least one balanced metric, (2) sensitivity, specificity, PPV and NPV, (3) seizure burden agreement, (4) multi-rater agreement using the statistical Turing test with Fleiss' $\kappa$, and (5) all of the above on a held-out validation set. This proposed framework provides an important prerequisite to clinical validation by enabling a thorough and honest appraisal of machine-learning or AI methods for neonatal seizure detection. 
\end{abstract}


\section{Introduction}
\label{sec:introduction}
Neonatal seizures represent a common neurological emergency in neonatal intensive care units (NICUs) \cite{b1, b2}, often caused by hypoxic-ischemic encephalopathy (HIE) and cerebrovascular injury, but may also arise from neonatal epileptic syndromes and structural brain malformations \cite{b1, b31}. Due to their association with adverse neurodevelopmental outcome, early and accurate detection is critical to minimise long-term consequences. However, seizures in neonates can be difficult to recognize, as they often occur without obvious clinical signs \cite{b3}. To overcome the limitations of clinical observation alone, continuous electroencephalographic (EEG) monitoring, which identifies seizures through characteristic evolving electrographic patterns, is used as the gold standard for seizure detection. Unfortunately, expertise in neonatal EEG interpretation is not always available \cite{b31, b32}. AI-based solutions have the potential to address this limitation through  expansion of monitoring capabilities to improve timely seizure identification \cite{b4, b5, b6, b7, b8, b9}. 

A recent clinical trial \cite{b10} demonstrated the potential of such automated systems to support real-world decision-making. Despite these advances, a critical limitation remains: the absence of complete and standardised evaluations. This makes it difficult to reliably assess and compare models, therefore increasing the risk of failure at the clinical investigation stage. Two particular challenges for neonatal seizure detection evaluation are pronounced class imbalance and the lack of a clearly defined ground truth. Since seizure annotations depend on expert interpretation of EEG, which may vary among clinicians, including multiple raters can help mitigate individual bias and better capture ambiguous cases. 

Evaluation practices in the field remain highly inconsistent, with different studies relying on a wide range of metrics; sample-based, event-based, and expert-level equivalence tests, with no consensus or guidelines on which to use \cite{b4, b5, b6, b7, b8, b9, b11, b12, b13, b14, b15, b16, b17, b18}. Although there have been previous efforts to standardise evaluation frameworks \cite{b10a, b10b}, no unified evaluation approach has been adopted. This inconsistency enables new methods to claim state-of-the-art performance based on selectively chosen metrics, while making it difficult for clinicians to interpret results or compare algorithms. The area under the receiver operating characteristic curve (AUC), for instance, is the most commonly reported metric \cite{b5, b6, b7, b8, b11, b12, b13, b14, b15, b17}, often used as the sole metric despite its known limitations \cite{b19, b20, b21}, while more robust alternatives are rarely applied. 
Recently, there has been a shift toward comparing AI methods directly to multiple human experts, with several studies claiming human--expert equivalence \cite{b9, b18, b22, b23}. However, due to the lack of standardised protocols, each study defines its own evaluation criteria and comparison strategies. These challenges underscore the urgent need for more objective and comparable assessment of model performance across studies that differ in dataset size, class distribution, and study setups.

The primary aim of this study is to establish a reliable and standardised framework for evaluating and comparing neonatal seizure detection models by systematically assessing commonly used performance metrics under varying class imbalance, numbers of raters, and inter-rater agreement (IRA). Establishing such an evaluation framework is a necessary first step toward clinical translation. An essential subsequent stage is to interpret model performance in terms of clinical decision-making, workflow integration, and acceptable false alarm rates, but this is out of scope for this work. We demonstrate that several commonly reported metrics are not well-adapted to the specific challenges of this domain and may yield overly optimistic results, particularly in the presence of extreme class imbalance. Developing and adopting standardised evaluation practices is therefore essential to ensure the reliability and comparability of algorithms and a fundamental requirement for clinical evaluation and integration of AI-based seizure detection tools. Although only neonatal EEG seizure detection is evaluated here, the proposed evaluation framework and findings may also be relevant to other EEG or time-series detection problems affected by annotation uncertainty and class imbalance.

\section{Methods}
\label{sec:methods}
We use seizure annotations from two neonatal EEG datasets: an open-access dataset referred to as the Helsinki dataset \cite{b17}, and a private dataset referred to as the Cork dataset \cite{b9, b24}. The Helsinki dataset includes annotations for short EEG recordings from 79 neonates; the Cork dataset contains annotations for long-duration EEG recordings from 51 neonates. Both datasets have annotations from three independent raters. Only annotations, not EEG records, are used in this study. To supplement these data, we develop a framework for generating synthetic data that mimic different characteristics of human annotation. This framework allows the simulation of multiple raters, multiple annotations per neonate, and controlled variations in disagreements, while maintaining a known ground truth—a characteristic absent in real-world datasets, where seizure annotations depend on subjective expert interpretation and no objective reference standard exists. Our motivation is to gain full control over the ground truth, including its prevalence, in order to systematically examine how different evaluation techniques behave under varying class imbalance, number of raters, and inter-rater agreement. In particular, we use this framework to study the impact of the frequency of different annotation error types on general performance metrics, and to assess the ability of human--expert equivalence tests to distinguish between rater groups with specific qualities or tendencies, such as well-calibrated raters versus systematic over- or under-raters. 

\begin{figure*}[!t]
\centerline{\includegraphics[width=\textwidth]{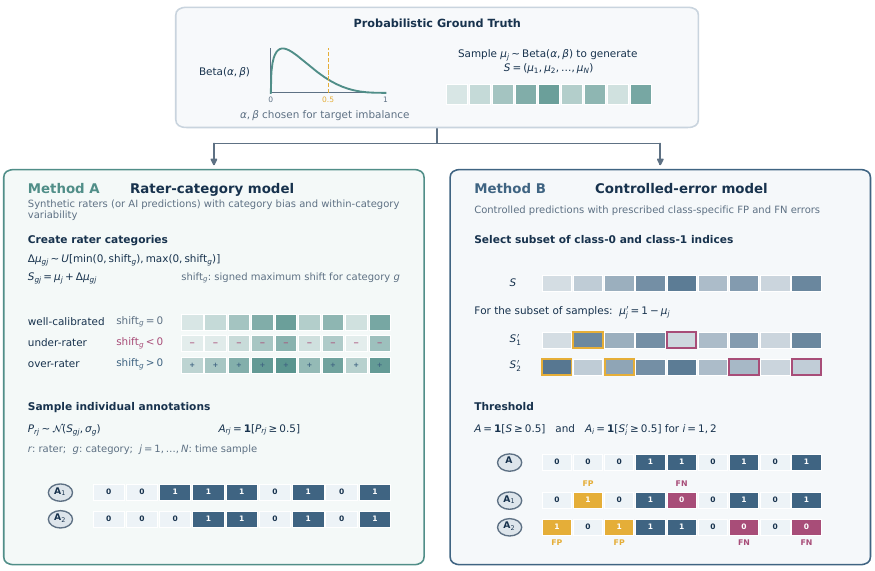}}%{METHODS.png}}
\caption{Overview of the Synthetic Annotation Framework (Methods A and B). A probabilistic ground truth is first generated by sampling seizure-occurrence probabilities from a beta distribution, whose parameters control seizure prevalence. Method A simulates multiple rater categories with different annotation tendencies, including well-calibrated raters (no shift), over-raters (positive shift), and under-raters (negative shift), by adding uniform shifts to the ground truth before sampling individual annotations. Method B introduces predefined FP and FN rates by flipping selected probabilities in the ground truth, preserving the probabilistic structure while altering class labels. Method A is used to evaluate consensus formation and human--expert equivalence tests, whereas Method B enables controlled evaluation of sample-based performance metrics. Both methods produce per-sample binary annotation sequences.}
\label{method1}
\end{figure*}

\subsection{Generating synthetic annotations}
\label{sec:gen_annot}
This framework generates per-sample seizure annotations, distinguishing between seizure (class 1) and non-seizure (class 0) independently for each sample (see Fig.~\ref{method1}). We first create a probabilistic ground truth, represented as a sequence of seizure occurrence probabilities $S = (\mu_1, \mu_2, ... \mu_N)$, where $N$ is the number of time samples in an EEG record. These values are sampled from a beta distribution, $\mathrm{Beta}(p, 1-p)$ (Fig.~\ref{method1}), where the parameter $p$ controls seizure prevalence. 

For a given ground truth $S$, we generate synthetic rater annotations using one of two methods. In Method A, we simulate different rater categories (Fig.~\ref{method1}) with different behavioural tendencies. This enables the simulation of multiple raters with controlled agreement within and between categories, making it ideal for testing expert-equivalence and consensus-based methods. Method B, on the other hand, provides fine control over the level of disagreement between two annotations,with known sensitivity and specificity values, supporting rigorous testing of general sample-based metrics under controlled conditions.

In Method A, we simulate three types of raters: well-calibrated raters who closely follow the ground truth, over-raters who detect more seizures, and under-raters who detect fewer seizures. Positive (negative) shift values are added to $S$ to create over-raters (under-raters), while well-calibrated raters retain the original $S$. For each rater category $g_i$, a per-sample shift vector is sampled from a uniform distribution: 
\begin{equation}
     \mathbf{\Delta}\mu^{g_i}  \sim U(min(0, \text{shift}_{g_i}), max(0, \text{shift}_{g_i})),\qquad i=1,...K\label{eq1}
\end{equation}
where $K$ is the number of rater categories, $\mathbf{\Delta\mu^{g_i}} = ( \Delta\mu_{1}^{g_i}, \Delta\mu_{2}^{g_i},...\Delta\mu_{N}^{g_i})$ represents the vector of shift values, and $\textrm{shift}_{g_i}$ is the maximum shift for category $g_i$, which is positive for over-raters, and negative for under-raters. This vector of shifts is added to the ground truth $S$ to create a group-specific base probability sequence $\mathbf{S^{g_i}}=S+\mathbf{\Delta\mu^{g_i}}$. The shift vector is sampled once per category, with a separate shift value for each time sample. Thus, \(S_j^{g_i}=\mu_j+\Delta\mu_j^{g_i}\) varies across time samples, while the full sequence \(S^{g_i}\) is shared by all raters in category \(g_i\).

Next, individual raters within each category are generated by introducing variability around this base probability sequence (Fig.~\ref{method1}). Specifically, for each rater and each sample index $j$, a probability is sampled from a Gaussian distribution:
\begin{equation}
     P^{g_i}_j \sim N(S_j^{g_i}, \sigma^{g_i}), j=1,...N\label{eq2}
\end{equation}
where $P_j^{g_i}$ represents the probability sampled from a Gaussian distribution for time step $j$ for the rater from category $g_i$, $S_j^{g_i}$ is the category-shifted probability at time step $j$, and $\sigma^{g_i}$ is a category-specific standard deviation that is fixed across time samples and raters within category $g_i$. A threshold of $0.5$ is applied to the resulting sequence $P_j^{g_i}$ to generate the accompanying binary annotation $A_j^{g_i}$ 
\begin{equation}
    \mathbf{A} = \begin{cases} 1 & \text{if} \quad \mathbf{P} \geq 0.5\\0 & \text{if} \quad \mathbf{P} < 0.5
    \end{cases}
    \label{eq3}
\end{equation}
where $A^{g_i}$ is the binary annotation for probabilities $\mathbf{P}^{g_i}$. Although this method does not explicitly model the typical source morphological patterns in the EEG that give rise to disagreement, by sampling model predictions from the same distribution as raters, we can evaluate whether the model behaves like human raters under controlled conditions.  This simplification may affect absolute measures of agreement between comparators but is unlikely to impact the relative reliability of different evaluations which is the focus of this work.

To test general performance metrics and IRA measures, we also developed Method B illustrated in Fig.~\ref{method1}. These experiments require precise control over the error rates for each class separately. With the ground truth created as before, we directly modify it to simulate predefined false positive (FP) and false negative (FN) rates, defined as the proportion of randomly selected samples to be altered in class 0 and class 1. Each altered probability $\mu_j$ is replaced with $1-\mu_{j}$, effectively flipping the class while preserving the probability structure. The resulting probability vectors are then binarized as described in Eq.~\ref{eq3}. Two variations of this method were used: Variation 1 introduces a fixed percentage of errors to both classes, proportional to vector length N, directly controlling sensitivity and specificity but altering class imbalance. Variation 2 controls sensitivity by flipping a fixed percentage of seizure samples to FNs, and then introduces an equal number of FPs in the non-seizure class (matching the FN count), thus maintaining the original class distribution. Table~\ref{tab:methods_table} summarises which method and variation underlies each experiment.

Both methods A and B provide a structured approach for generating sample-based  annotations, particularly well suited for evaluating metrics that operate at the time-sample level. Since samples are generated independently, the framework does not reproduce physiologically realistic seizure-event durations, temporal clustering, inter-event intervals, or event boundaries. It is therefore used to evaluate sample-based metrics and is not applied to event-based metrics, whose interpretation depends on realistic temporal event structure and explicitly defined event-matching and post-processing rules. However, it enables a controlled comparison between rater annotations and model predictions, allowing a more precise assessment of model performance in seizure detection. To enable reproducible research, the computer code is freely available\footnote{\url{https://github.com/jovanak1/annotation-generation-and-expert-level-testing/tree/189ee47}}.

%\hl{Method A enables the simulation of multiple raters with controlled agreement within and between categories, making it ideal for testing expert-equivalence and consensus-based methods. By generating synthetic model predictions by sampling from the same distribution as raters, we can evaluate whether the model behaves like human raters under controlled conditions.} Conversely, by modifying the distribution, we can assess deviations in model performance and determine whether it departs from expert-like decision-making. This provides a systematic way to test methods designed to establish expert-equivalence in seizure detection. Rater and AI bias in this framework is modelled only as an overall uniformly distributed tendency, whereas individual annotator behaviour is generated using Gaussian sampling to capture subject-level variability. It does not explicitly capture disagreement arising from morphological patterns in the EEG that may resemble seizure activity. However, because both simulated raters and the simulated AI predictions are generated under the same assumptions, this simplification is unlikely to affect their relative comparison, although it may influence absolute agreement values.

%Method B, on the other hand, provides fine control over the level of disagreement between two annotations, with known sensitivity and specificity values, supporting rigorous testing of general sample-based metrics under controlled conditions. \textcolor{green!60!black}{Table~\ref{tab:methods_table} summarises which method and variation underlies each experiment.}

\begin{table}
\centering
\caption{Use of the synthetic annotation methods across the experiments. Method A varies who is annotating: it produces any number of raters whose agreement emerges from the category shift and within-category variability, and one of them can be designated as the AI predictions. Method B varies how many errors are made: it produces a single prediction with exactly prescribed class-specific error rates against a known reference.}
\label{tab:methods_table}
\setlength{\tabcolsep}{3pt}
\begin{tabular}{|l|l|l|}
\hline
\textbf{Experiment} & \textbf{Method}\\% & \textbf{Varied} \\
\hline
Sample-based metrics (Section~\ref{sec:sec31}) & B, Variation 1\\% & Imbalance 1:1--50:1, fixed 10\% error \\
\hline
Consensus strategies (Section~\ref{sec:sec32}) & A \\%& 3--15 raters, Fleiss' $\kappa$ 0.5--0.9 \\
\hline
Real-data-informed comparison (Section~\ref{sec:sec32}) & A \\%& Three-rater configurations informed by broad Helsinki and Cork annotation characteristics \\
\hline
Equivalence tests (Section~\ref{sec:sec33}) & A \\%& Expert:non-expert 1--29, bias type, imbalance \\
\hline
IRA measures (Appendix~\ref{app:app1}) & B, Variation 2 \\%& Disagreement 0--100\% at fixed imbalance \\
\hline
\end{tabular}
\end{table}

\subsection{General Performance Metrics}
A frequently used method for evaluating seizure detection models is to compute sample-based performance metrics comparing the annotation (or consensus annotation in the case of multiple raters) to the AI predictions. Neonatal EEG data are typically highly imbalanced; for example, a non-seizure to seizure ratio of approximately 50:1 was observed in a large dataset comprising over 50,000 per-channel hours of annotated EEG recordings \cite{b9}. Pronounced class imbalance is also expected in real clinical settings and so performance should be evaluated in this context. As seizures represent the minority class, metrics should specifically reflect how well the model identifies true positives (TPs) and avoids FPs, the latter of which is often ignored to artificially boost performance assessment. 

\subsubsection{Sample-Based Metrics}%\leavevmode\\

The most commonly reported metric for neonatal seizure detection is the AUC. It is often the only reported metric, despite its many limitations \cite{b19, b20, b21}. In addition to AUC, some studies report sensitivity and specificity, while others include positive predictive value (PPV) and negative predictive value (NPV) \cite{b6}. Sensitivity and specificity reflect the percentage of correctly classified positive and negative instances, respectively. When these percentages are fixed, the metrics remain unchanged across different class imbalance scenarios, even though the actual proportions of FPs/FNs vary drastically, making them insensitive to such shifts \cite{b25}. In contrast, PPV and NPV are sensitive to the described class imbalance in datasets \cite{b26}, but they are insensitive when the proportion of TPs/FPs or TNs/FNs, respectively, remain unchanged \cite{b26}. Although all four metrics provide complementary information on model performance, they are rarely reported together in ML-based EEG studies, and reporting only two of them (e.g., sensitivity/specificity) risks overlooking critical evaluation insights.

Comparing models across four separate metrics can be challenging. A single, comprehensive metric that integrates all four elements of the confusion matrix into one value would simplify comparisons while accounting for class imbalance. This ideal evaluation metric should be robust to class imbalance, exhibit a linear or nearly linear progression across a performance range, remain stable regardless of dataset size, be easily interpretable, and deliver consistent results across various datasets. One metric that accounts for all four elements of the confusion matrix is the Matthews correlation coefficient (MCC) \cite{b20}. MCC evaluates performance by producing a high score only when TP and TN are high and FP and FN are low \cite{b27}. We also explore how MCC performs in relation to other metrics and include its continuous-valued equivalent, Pearson’s correlation coefficient (PCC). 

Seizure burden is a sample-based, clinically meaningful metric reflecting seizure severity. Several neonatal studies have demonstrated a strong correlation between higher seizure burden and more adverse neurodevelopmental outcomes (see e.g. \cite{b28}). Because of this established clinical relevance, seizure burden should be estimated and reported alongside other metrics. It is defined as the cumulative seizure duration over a given period of time, expressed in minutes per hour \cite{b9, b29}, while total seizure burden represents the cumulative seizure duration over the entire recording and is expressed in minutes. To evaluate how well an AI model estimates seizure burden, correlation (using PCC) can be calculated between model-estimated and reference seizure burden across corresponding hourly windows. A high correlation indicates that the model tracks temporal variations in seizure burden, even if individual seizure events are not perfectly aligned; however, it does not by itself establish close numerical agreement between the model and reference burden estimates. As a complementary summary of numerical agreement, the absolute difference between model-estimated and reference mean seizure burden over each neonate’s recording could be reported in min/h and summarised across neonates using the median and interquartile range.

\subsubsection{Event-Based Metrics}%\leavevmode\\
Event-based metrics are also used to evaluate seizure detection models, as seizures represent continuous episodes rather than independent time samples. Among the most widely used are event-based sensitivity and the false detection rate per hour (FD/h) \cite{b4, b5, b14}. These metrics are often used in the neonatal EEG literature because they express performance in terms that map directly onto clinical utility, and they remain valuable as complementary measures when reported together with sample-based metrics and seizure burden.

Event-based sensitivity measures the proportion of true seizure events correctly detected by the model, where an event is considered detected if there is any overlap between the predicted and annotated seizure \cite{b4}. While this metric provides a clinically intuitive measure of detection performance, it has important limitations when reported without accompanying sample-based metrics. Under an any-overlap criterion, a model predicting seizure continuously would achieve perfect event-based sensitivity while potentially producing only one long false-detection event, resulting in a deceptively low FD/h despite being clinically unusable. This combination of event-based sensitivity and FD/h therefore rewards predicting long-duration seizure events, as FD/h is independent of event duration.

\subsection{Consensus Types}
Seizure annotations are often performed by multiple raters \cite{b5, b16, b17, b18} with their consensus serving as the ground truth. Unanimous consensus \cite{b5, b7, b8, b9, b16, b17, b18} retains only signal segments where all raters agree, discarding the rest. While this approach ensures high-confidence annotations, it reduces dataset size and may exclude informative segments, potentially impacting both analysis and model performance. Majority consensus \cite{b9, b22} ensures that no data are discarded while reflecting majority agreement among raters. Retaining ambiguous data can be useful for representing uncertainty in the ground truth, but it makes model error analysis less clear in cases with weak majority agreement (e.g. 6 out of 10 raters agree). A third option is consensus through joint review \cite{b5} where differing opinions are reconciled through collective re-evaluation of the EEG. While appealing, it is labour-intensive and can be influenced by a small number of raters in the panel without always representing the views of the majority. Methods for single annotations can be applied to the joint-review consensus.

\subsection{Human--Expert Equivalence Testing}
\label{sec:exp_tests}
%\vspace{-3mm} % above figure
\begin{figure*}[!t]
\centerline{\includegraphics[width=\textwidth]{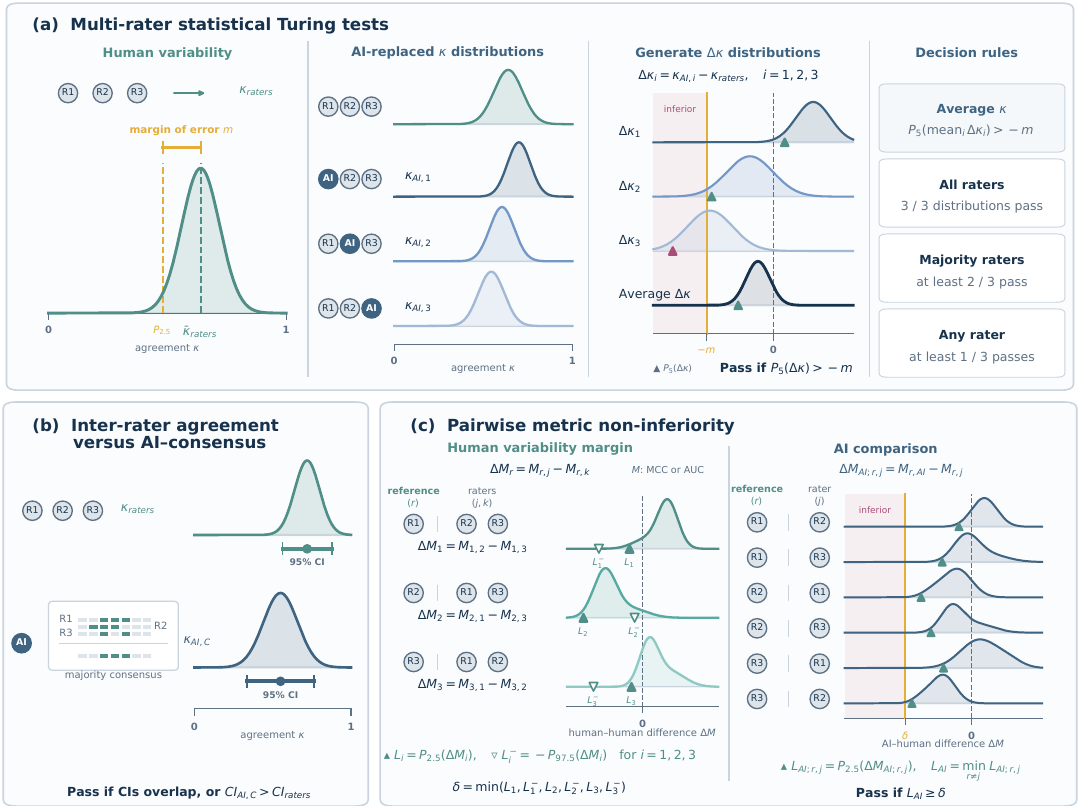}}%{expert-level-tests.png}}
    \caption{Overview of different human--expert equivalence tests. (a) The multi-rater agreement statistical Turing test replaces each rater with the AI and measures the resulting change in IRA ($\Delta \kappa_i=\kappa_{\text{AI}}-\kappa_{\text{raters}}$), assessing whether the AI can substitute for a human expert. If the 5th percentile of $\Delta \kappa$ is higher than the margin, the model is considered sufficiently reliable. Fleiss' $\kappa$ is used here as the IRA metric, but other agreement measures (e.g., Gwet's AC1, Krippendorff's $\alpha$) can be substituted within the same test structure. (b) The IRA vs. AI-Consensus Agreement test compares IRA among human raters ($\kappa_{\text{raters}}$) and AI-majority human consensus agreement ($\kappa_{\text{AI, Consensus}}$), using bootstrapping to estimate $95\%$ CIs. (c) In the Pairwise Metric Statistical Non-inferiority Test, each rater serves as the reference to compute pairwise metrics M (e.g., MCC) with others and the AI. Differences between human--human scores are used to define non-inferiority margins (e.g., $\text{MCC}_{\text{R1,R2}} - \text{MCC}_{\text{R1,R3}}$). For each reference rater, the human--human difference distribution is used to define both bounds of the non-inferiority margin: the lower bound as its 2.5th percentile, and the upper bound via the reflected 97.5th percentile ($-P_{97.5}(\Delta M) = P_{2.5}(-\Delta M)$), since the reflected distribution is equivalent to the original under sign inversion and is therefore not shown separately. The non-inferiority margin $\delta$ is defined as the smallest of the human--human bounds ($L_1, L_1^-, L_2, L_2^-, L_3, L_3^-$) across all reference raters. Similarly, for the AI, $L_{\text{AI};r,j} = P_{2.5}(\Delta M_{\text{AI};r,j})$ is computed for each reference-rater pair, and the smallest of these, $L_{\text{AI}} = \min_{r\neq j} L_{\text{AI};r,j}$, is compared against $\delta$: the AI passes if $L_{\text{AI}} \geq \delta$.}
\label{expertlevel}
\end{figure*}

While general performance metrics are useful for model comparison, they rely on consensus annotations that fail to capture inter-rater variability. This is particularly relevant in seizure detection, where no objective ground truth exists. To address this, human--expert equivalence tests evaluate whether AI performance falls within the expected range of variability observed among raters. Such methods provide a more clinically meaningful benchmark; if a model performs at the level of experts, it can be considered a viable decision-support tool for seizure detection. This makes human--expert equivalence testing an essential validation step for translating automated seizure detection systems into real-world clinical practice. Unfortunately, there is no accepted standard for these tests, with several variants reported in the literature. These tests can be categorized into the following groups, with illustrations in Fig.~\ref{expertlevel}:

    \textbf{1. Multi-Rater Agreement Statistical Turing Tests} -  The original version of this test \cite{b9} determines whether a model passes by checking if the 95\% CI of the difference between human-only and AI-substituted IRA includes zero for at least one substitution. We reformulate this as a non-inferiority test that instead assesses whether the model falls within the expected range of inter-rater variability, as follows. First, the IRA distribution is computed among human raters using bootstrap resampling. The margin-of-error of this distribution, calculated as the mean minus the 2.5th percentile, is used to generate a lower-threshold bound, m. Next, $\Delta \kappa$ values, calculated as $\Delta \kappa_i=\kappa_{\text{AI}}-\kappa_{\text{raters}}$, are computed by iteratively substituting the AI for each expert and calculating the mean $\Delta \kappa$ for each bootstrap sample, producing a bootstrap distribution of $\Delta \kappa$ values. The AI is considered equivalent to expert-level performance if the lower 5th percentile of the $\Delta \kappa$ distribution exceeds the lower threshold bound -m defined from the expert-only distribution. We used Fleiss’ $\kappa$ as the IRA metric in four test variations:
    %This approach evaluates whether a model can sufficiently replace a human rater. IRA is first computed using only human annotations, and then recomputed after replacing one human with the model. The $95\%$ CI of the difference between the two IRAs is calculated via bootstrap resampling. If the $95\%$ CI includes 0 for at least one substitution, the model passes the test \cite{b18}. We used Fleiss’ $\kappa$ as an IRA metric in four test variations:
    \renewcommand{\labelenumi}{\theenumi)}
    \begin{enumerate}
        \item \textit{Average $\kappa$} – Perform at the level of average agreement. The $\Delta \kappa$ distribution is computed for the mean difference between all AI-replaced configurations and the human-only IRA \cite{b9}.
        
        \item \textit{All raters} - Outperform all raters. Each $\Delta\kappa_i$ distribution must independently satisfy the pass criterion: $P_5(\Delta\kappa_1) > -m$, $P_5(\Delta\kappa_2) > -m$, and $P_5(\Delta\kappa_3) > -m$, assuming three raters in this example.
        
        \item \textit{Majority raters} – Outperform the majority of raters. 
        
        \item \textit{Any rater} – Outperform at least one rater \cite{b18}.% (any CI excludes 0) \cite{b18}.
    \end{enumerate}
    with 1) also tested using Gwet’s AC1, {\it Average AC1}.

    \textbf{2. IRA versus AI--Consensus Agreement Tests} - This approach compares the agreement observed among human raters to the agreement between the AI and the majority human consensus, using bootstrap resampling to estimate 95\% confidence intervals for both. Specifically, IRA among human raters, $\kappa_{\text{raters}}$, is computed via bootstrap resampling, yielding a confidence interval, $\textrm{CI}_{\text{raters}}$. Similarly, the agreement between the AI and the majority consensus of human raters, $\kappa_{\text{AI,Consensus}}$, is computed via bootstrap resampling, yielding $\textrm{CI}_{\text{AI,Consensus}}$. The AI is considered to fall within the human variability range, and therefore passes the test, if $\textrm{CI}_{\text{raters}}$ and $\textrm{CI}_{\text{AI,Consensus}}$ overlap, or if $\textrm{CI}_{\text{AI,Consensus}} > \textrm{CI}_{\text{raters}}$ \cite{b22}. Two variants were implemented using:
    %\textbf{2. IRA vs. AI-Consensus Agreement Tests} - \hl{These methods compare the agreement among human raters, $\textrm{IRA}_{raters}$, to agreement between the model and the consensus annotation, $\textrm{IRA}_{AI, Consensus}$), using an IRA measure}. If the CIs of $\textrm{IRA}_{raters}$, $\textrm{CI}_{raters}$, and AI-consensus agreements, $\textrm{CI}_{AI, Consensus}$, overlap, or if the $\textrm{CI}_{AI, Consensus} > \textrm{CI}_{raters}$, the model is considered within the human variability range and passes the test \cite{b22}. \hl{Two variants were implemented using:}
    %a. Gwet's AC1 as the IRA measure} \cite{b22} ({\it IRA vs. AI-Consensus AC1}), b. Fleiss' $\kappa$ as the IRA measure ({\it IRA vs. AI-Consensus $\kappa$}).
    \renewcommand{\labelenumi}{\theenumi)}
    \begin{enumerate}
        \item   \textit{IRA vs. AI-Consensus AC1} - Uses Gwet's AC1 as the IRA measure \cite{b22}.
        \item   \textit{IRA vs. AI-Consensus $\kappa$} - Uses Fleiss' $\kappa$ as the IRA measure.
    \end{enumerate}

    \textbf{3. Pairwise Metric Statistical Non-inferiority Tests} - This approach compares AI performance to human raters using a predefined performance metric (e.g., MCC, AUC). In the absence of ground truth, each human rater is alternately treated as the reference, and the remaining raters and the AI are compared against this reference in each bootstrap iteration. Pairwise human--human differences define the empirical performance range, from which $95\%$ CIs are computed to establish non-inferiority margins. The AI is considered non-inferior if the lower bound of its CI exceeds the non-inferiority margin. Fig.~\ref{expertlevel} illustrates this procedure for three raters. We implemented this test using either AUC or MCC as the predefined performance metric: 
    %a. AUC ({\it Pairwise AUC}), b. MCC ({\it Pairwise MCC}).
    \renewcommand{\labelenumi}{\theenumi)}
    \begin{enumerate}
        \item   \textit{Pairwise AUC}
        \item   \textit{Pairwise MCC}
    \end{enumerate}

% \par je za drugi red
\begin{table}
\centering
\caption{Overview of the four dataset groups (D1–-D4). Class distribution is the non-seizure:seizure ratio. Non-expert category describes how simulated non-expert raters deviate from the ground truth: systematic over- and under-annotation (D1, D2) or errors without consistent direction (D3, D4). Each group contains 29 datasets of 30 raters, with the number of experts varied from 1 to 29.}
\label{table1}
\setlength{\tabcolsep}{3pt}
\begin{tabular}{|l|l|l|}%{|p{25pt}|p{75pt}|p{115pt}|}
\hline
\textbf{Groups} & \textbf{Class Distribution} & \textbf{Non-expert category} \\
\hline
D1 & Balanced 1:1 & over- and under-raters \\
\hline
D2 & Imbalanced 25:1 & over- and under-raters \\
\hline
D3 & Balanced 1:1 & directionless annotation errors \\
\hline
D4 & Imbalanced 25:1 & directionless annotation errors \\
\hline
\end{tabular}
\end{table}

\subsubsection{Evaluating Human--Expert Equivalence Tests}%\leavevmode\\
Although various human--expert equivalence tests have been proposed in the literature, there is no consistency in their use and it remains unclear which tests satisfy these criteria. Given the significance of the claim of expert-level equivalence, it is important to establish the validity of these tests. The evaluation of human–expert equivalence tests was based on a combination of qualitative and quantitative criteria.

    \textbf{Qualitative criteria:} An ideal expert-level test should meet the following qualitative criteria:
    \begin{enumerate}
    \item \textbf{Flexible to the number of raters} – Allows for any number of raters and becomes more reliable with more raters.

    \item \textbf{Robustness to class imbalance} – Maintains reliability across varying non-seizure to seizure class distributions.

    \item \textbf{Robustness to outliers} – Remains stable in the presence of outlier raters.

    \item \textbf{Missing data resilience} – Ideally, it should be capable of handling missing annotations without compromising validity. For example, in very large datasets, it may be impractical to have all raters annotate every EEG record or segment, or different raters may annotate different subsets.
    \end{enumerate}

        \textbf{Quantitative criteria:} We also frame the evaluation of human--expert equivalence tests as a binary classification task, aiming to distinguish experts (positive class) from non-experts (negative class) based on test pass/fail outcomes. For each synthetic dataset, we compute classification accuracy, assuming that experts should pass and non-experts should not. To account for varying expert prevalence, we introduce a Weighted Accuracy ($\mathcal{A}_W$), which assigns greater importance to datasets with higher proportions of experts, where misclassifying a non-expert as an expert is more critical. A higher proportion of experts also reflects higher-quality ground truth and stronger consensus. For each dataset $i$, let $a_i$ denote the classification accuracy and $e_i$ the number of experts. We define the weight $\omega _ i=e_i/\sum_{j=1}^N{e_j}$, where $N = 29$ is the total number of datasets compared, and $\sum_i\omega _i=1$. The $\mathcal{A}_W$ is then computed as:
        \begin{equation}
           \mathcal{A}_W = \sum_{i=1}^{N}\omega _i a_i \label{weighted_acc}
        \end{equation}
        The result is a single, interpretable score that prioritizes accurate classification in expert-dense scenarios while down-weighting lower-quality ones. 

        To assess how the tests behave under different annotation conditions, we conducted experiments on multiple synthetic datasets generated using Method A (Section \ref{sec:gen_annot}). Four dataset groups were created with varying class distributions (balanced vs. unbalanced class distribution) and annotation characteristics (over-rater non-experts, under-rater non-experts, and mixed-error non-experts without bias). Table~\ref{table1} summarises these dataset groups (D1--D4). Each dataset group includes 29 datasets, with 30 raters per dataset. The number of experts was systematically varied from 1 to 29, with the remaining raters designated as non-experts. Here, the term non-expert refers to simulated inaccurate raters rather than qualified reviewers with less experience. Their annotations are intentionally generated as imperfect approximations of the expert annotations. Each rater was evaluated against the other 29, and pass/fail outcomes were tracked by rater type across different expert/non-expert ratios (116 datasets in total). This setup allowed for a detailed analysis of how the tests respond to annotation experience, rater bias, and class imbalance. %\textcolor{blue}{It is worth noting that rater and AI bias in this framework is modelled only as an overall uniformly distributed tendency, whereas individual annotator behaviour is generated using Gaussian sampling to capture subject-level variability. This framework does not explicitly capture disagreement arising from morphological patterns in the EEG that may resemble seizure activity.}

\section{Results}

\subsection{General Performance Metrics}
\label{sec:sec31}

\begin{figure}[!bt]
\centerline{\includegraphics[width=\columnwidth]{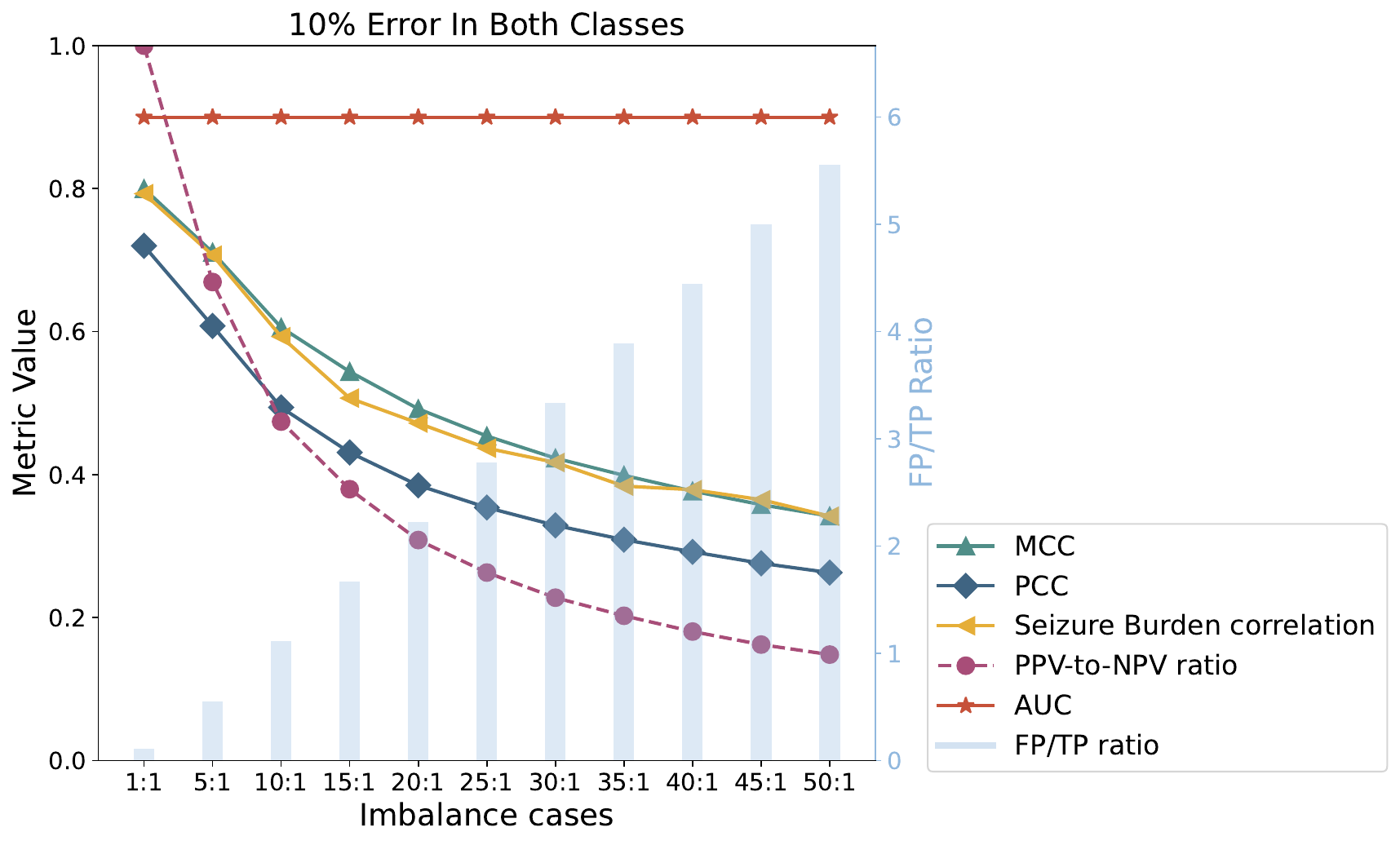}}
\caption{Influence of class imbalance on sample-based performance metrics. Synthetic predictions were generated with Method B (Variation 1) at a fixed error rate corresponding to $90\%$ sensitivity and $90\%$ specificity while progressively increasing the non-seizure:seizure ratio from balanced to highly imbalanced. Increasing class imbalance increases the FP/TP ratio and decreases PPV. Unlike MCC, PCC and seizure burden correlation, the AUC remains constant because it depends only on sensitivity and specificity, illustrating its insensitivity to class imbalance.}
\label{auc}
\end{figure}

To test the robustness of metrics to class imbalance, we used Method B - Variation 1 (Section~\ref{sec:gen_annot}, Table~\ref{tab:methods_table}) to generate synthetic predictions with a fixed $10\%$ error ($90\%$ sensitivity and specificity), relative to the ground truth, while progressively increasing class imbalance from balanced to 50:1. Increasing the class imbalance leads to an increased FP/TP ratio. As shown in Fig.~\ref{auc}, AUC remained high across all imbalance levels, despite increasing FPs and reduced PPV. For example, the AUC remained at 0.9 whether the FP/TP ratio is below 1 or above 5, due to its dependence only on sensitivity and specificity. In contrast, we found MCC and PCC more effectively captured performance degradation as the ratio of FP/TP increases. Crucially, the seizure burden correlation, a clinically meaningful outcome measure, follows a similar decreasing trend with increasing  FP/TP ratios, highlighting the real-world consequences of excessive FPs.

\subsection{Comparison of Consensus Types}
\label{sec:sec32}
\begin{figure}[!bt]
\centerline{\includegraphics[width=\columnwidth]{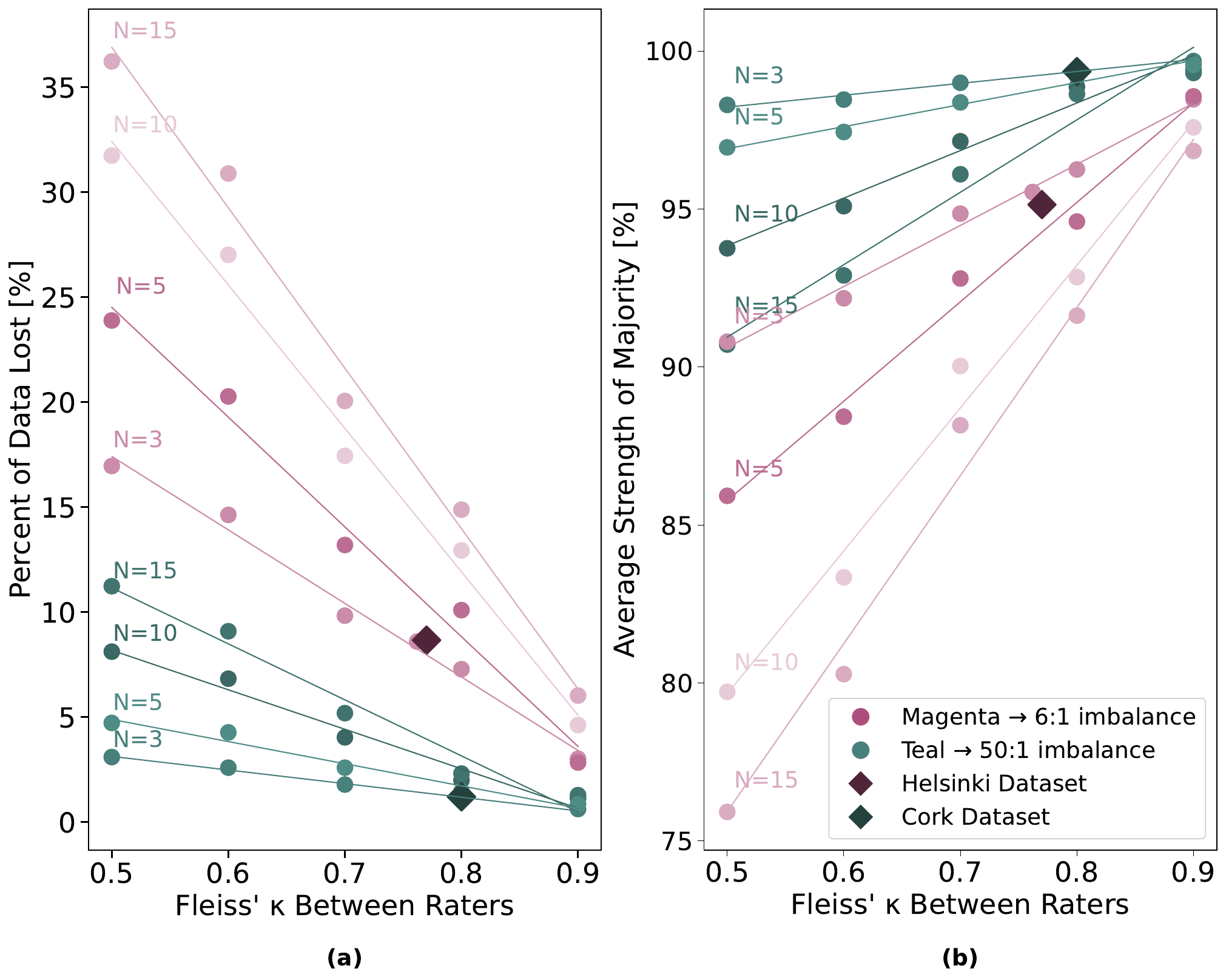}}
\caption{Effect of inter-rater agreement and number of raters on consensus annotation. (a) Percentage of samples discarded when unanimous consensus is used. (b) Average proportion of raters supporting the majority decision (majority strength). Results are shown for varying numbers of raters and levels of Fleiss' $\kappa$. Data points from real-world datasets: the Helsinki dataset (3 raters, Fleiss' $\kappa$ = 0.77, imbalance $\approx$ 6:1) \cite{b17} and the Cork dataset (3 raters, $\kappa$ = 0.80, imbalance $\approx$ 50:1) \cite{b9}, are included for comparison. The figure illustrates that unanimous consensus increasingly discards data as the number of raters grows, while majority consensus becomes progressively weaker.}
\label{consensus_results}
\end{figure}

To evaluate how rater count and IRA influence consensus selection, we generated two synthetic annotation sets using Method A (Section \ref{sec:gen_annot}), approximating the class imbalance observed in two real-world datasets: the Helsinki dataset \cite{b17} (3 raters, Fleiss’ $\kappa = 0.77$, class imbalance $\sim 6:1$) and the Cork dataset \cite{b9, b24} (3 raters, $\kappa = 0.80$, imbalance $\sim 50:1$). The number of raters was varied from 3 to 15, with Fleiss’ $\kappa$ ranging from $0.5$ to $0.9$. Fig.~\ref{consensus_results} illustrates how the percentage of data discarded increases with decreasing agreement among raters. Additionally, as the number of raters increases, the average strength of majority consensus decreases (Fig.~\ref{consensus_results}b), even when Fleiss’ $\kappa$ among raters remains constant. 
Both real-world cases are included in the plots (Fig.~\ref{consensus_results}a and \ref{consensus_results}b), and their alignment with the synthetic results supports the validity of our modelling approach for generating representative annotation scenarios.

\begin{figure*}[!bt]
\centerline{\includegraphics[width=0.8\textwidth]{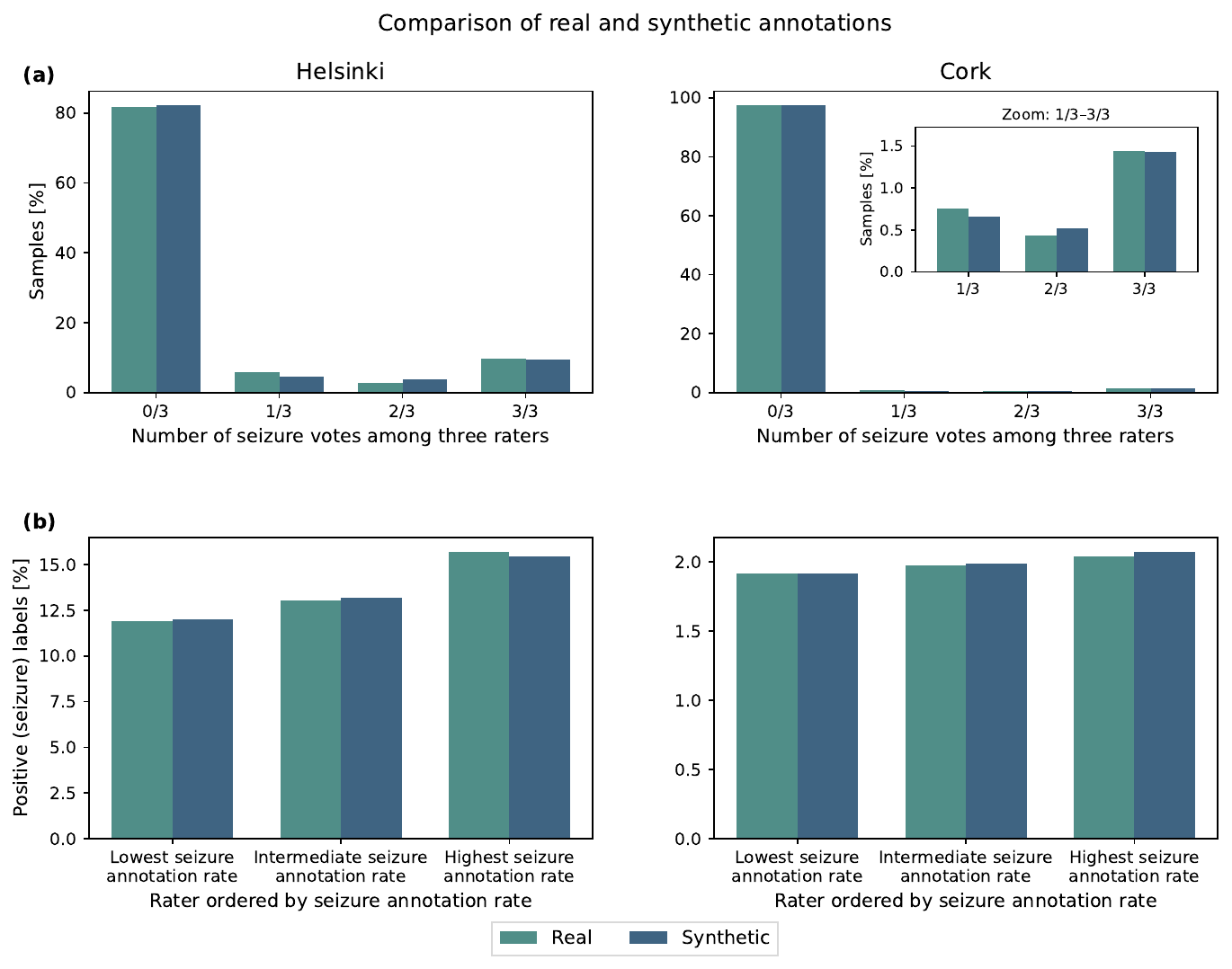}}
\caption{Comparison of real and matched synthetic three-rater annotation characteristics.  For each dataset, Method A was used to construct a configuration with one unshifted rater, one under-rater, and one over-rater, informed by the number of raters, their relative positive-label tendencies, and overall Fleiss' $\kappa$,  in the Helsinki (79 neonates) and Cork (51 neonates) datasets. (a) Distribution of samples receiving seizure labels from 0, 1, 2, or all 3 raters. (b) Proportion of positive seizure labels for the three raters, ordered independently within each dataset from the lowest to the highest seizure annotation rate. The comparison assesses whether Method A can reproduce broad aggregate annotation patterns and is not an independent external validation.}
\label{synthetic_annotations_compared_to_real}
\end{figure*}

To examine whether Method A can produce rater-specific annotation tendencies resembling those observed in real multi-rater data, we constructed a simple three-rater configuration informed by broad characteristics of the Helsinki and Cork datasets. Specifically, we considered the number of raters, their relative tendencies to annotate less or more seizure activity, and the overall Fleiss’ $\kappa$. Accordingly, each synthetic configuration comprised of one unshifted rater, one under-rater, and one over-rater. Method A generates the annotations through its rater-category shift and variability parameters; these parameters were not fitted to individual neonates, seizure counts, event characteristics, or the detailed annotation distributions of either dataset.
The resulting synthetic annotations reproduced the ordering of lower, intermediate, and higher positive-label rates observed among the real raters (Fig.~\ref{synthetic_annotations_compared_to_real}b), showing that Method A can represent broad systematic differences in the amount of seizure activity annotated.
As an additional descriptive comparison, we examined the distribution of samples receiving 0, 1, 2, or 3 seizure votes (Fig.~\ref{synthetic_annotations_compared_to_real}a). This distribution was not used to define the synthetic configurations. The dominant 0/3 and 3/3 categories were similar in the real and synthetic annotations for both datasets (Helsinki: 81.7\% vs 82.1\% and 9.8\% vs 9.5\%; Cork: 97.4\% vs 97.4\% and 1.4\% vs 1.4\%), whereas the less frequent Helsinki 1/3 and 2/3 categories differed more (5.8\% and 2.8\% real vs 4.6\% and 3.7\% synthetic). These results show that a configuration based on a small set of broad annotation characteristics can generate similar aggregate patterns, without implying exact reproduction of the datasets or independent external validation.

\subsection{Human--Expert Equivalence Testing}
\label{sec:sec33}
Multiple expert-equivalence test variants were evaluated using four synthetic dataset groups (D1-–D4) designed to systematically vary annotation bias, rater composition, and class imbalance (see Section \ref{sec:exp_tests} for details).  

\subsubsection{Quantitative evaluation}%\leavevmode\\
Table~\ref{tab:results_table} shows $\mathcal{A}_W$ scores across all dataset groups and test variants. The $\mathcal{A}_W$ measures the correct identification of the experts versus the non-experts in the synthetic datasets. Across the four controlled synthetic dataset groups, {\it Average $\kappa$} achieved the highest weighted accuracy among the evaluated procedures ($\mathcal{A}_W: 0.964 - 0.993$), indicating strong discrimination between expert and non-expert raters across the class distributions and rater biases examined. In contrast, {\it Average AC1} performed well on balanced datasets (D1, D3), but decreased in imbalanced ones (D2, D4), showing AC1’s sensitivity to class imbalance.

The {\it Any rater} test produced the lowest scores across all dataset groups ($0.658-0.662$), comparable to a baseline $\mathcal{A}_W$ value of $0.656$ where all raters pass regardless of expertise level, indicating that, in these experiments, the test did not discriminate between expert and non-expert raters. Similarly low results were observed for {\it Pairwise MCC} and {\it AUC}, both of which yielded $\mathcal{A}_W \sim 0.658 – 0.690$. The remaining tests showed moderate performance, with $\mathcal{A}_W$ values relatively consistent across different levels of class imbalance and rater compositions.

To ensure that this advantage is not an artefact of the $\mathcal{A}_W$ weighting scheme - which by design emphasises performance in expert-dense configurations - we additionally conducted a qualitative, unweighted analysis (see Appendix \ref{app:app2} Fig.~\ref{D1}) examining raw pass/fail rates across the full range of expert proportions. This complementary unweighted analysis shows that the {\it Average $\kappa$} test exhibits the most consistent behaviour across the entire range of expert compositions, not only in expert-dense conditions, indicating that its advantage is not driven by the $\mathcal{A}_W$ weighting alone.

\begin{table}[t]
\centering
\caption{Summary of weighted accuracy $\mathcal{A}_W$ to distinguish expert from non-expert raters. D1, D2, etc. are specific datasets defined in Table~\ref{table1}. Higher values indicate better discrimination between experts and non-experts. The best-performing method for each dataset group is highlighted in bold. For reference, a test passing all raters regardless of expertise yields $\mathcal{A}_W= 0.656$.}
\renewcommand{\arraystretch}{1.3}
\begin{tabular}{|c|c|c|c|c|c|}
\hline
\textbf{Test Type} & \textbf{Metric/Criterion} & 
\textbf{\makecell{D1\\ (1:1)}} &% \par \textbf{(1:1, 3)} & 
\textbf{\makecell{D2\\ (25:1)}} &%\par \textbf{(50:1, 3)} & 
\textbf{\makecell{D3\\ (1:1)}} &%\par \textbf{(1:1, 2)} & 
\textbf{\makecell{D4\\ (25:1)}} \\%\par \textbf{(50:1, 2)}  \\
\hline

\multirow{6}{*}{\makecell{Multi-Rater \\ Turing Test}} 
 & \makecell{\it Average $\kappa$} & \textbf{0.993*} & \textbf{0.964*} & \textbf{0.991*} & \textbf{0.987*} \\
\cline{2-6}
 & \makecell{\it Average AC1} & \textbf{0.993*} & 0.848 & \textbf{0.991*} & 0.802 \\
 \cline{2-6}
 & \makecell{\it Majority raters} & 0.882 & 0.926 & 0.849 & 0.853 \\
 \cline{2-6}
 & \makecell{\it Any rater} & 0.662 & 0.658 & 0.658 & 0.658 \\
 \cline{2-6}
 & \makecell{\it All raters} & 0.746 & 0.817 & 0.765 & 0.731 \\
\hline

\multirow{2}{*}{\makecell{IRA vs. \\AI-Consensus}}
 & {\it Gwet’s AC1} & 0.872 & 0.847 & 0.854 & 0.854 \\
 \cline{2-6}
 & {\it Fleiss’ $\kappa$} & 0.884 & 0.842 & 0.854 & 0.854 \\
\hline

\multirow{2}{*}{Pairwise Test}
 & {\it MCC} & 0.660 & 0.690 & 0.660 & 0.658 \\
 \cline{2-6}
 & {\it AUC} & 0.660 & 0.690 & 0.660 & 0.658 \\
\hline

\end{tabular}

\vspace{1mm}
\footnotesize
* Indicates the best result on dataset group.
\label{tab:results_table}
\end{table}

\subsubsection{Qualitative evaluation}%\leavevmode\\
\textbf{Flexibility to Number of Raters:} The {\it Average $\kappa$} test showed the most consistent behaviour; experts always passed, while non-experts were increasingly rejected as the number of experts increased. Moderate performance was observed in the {\it Majority raters}, as well as in both {\it IRA vs. AI-Consensus AC1} and {\it IRA vs. AI-Consensus $\kappa$} tests. These improved with more experts but with less precision than the {\it Average $\kappa$} test. Poor performance was found in the {\it Any rater} test, which failed to reject non-experts regardless of the expert count, and both {\it Pairwise MCC} and {\it AUC} tests, which rejected a small fraction of non-experts. In contrast, the {\it All raters} test was overly strict: although it correctly rejected all non-experts in every case, it rejected experts too frequently.

\textbf{Robustness to Class Imbalance:} Comparing results across dataset groups D1 vs. D2 and D3 vs. D4 highlights sensitivity to class imbalance. All tests (except {\it All raters}) consistently allowed experts to pass (see Appendix~\ref{app:app2}, Fig.~\ref{D1}), with differences mainly reflecting how strictly non-experts were rejected. The {\it Average AC1} achieved the highest $\mathcal{A}_W$ in balanced datasets (0.993 in D1, 0.991 in D3), but dropped in imbalanced ones (0.848 in D2, 0.802 in D4), confirming AC1’s sensitivity to skewed class distributions. Appendix \ref{app:app1} includes more information on AC1's sensitivity to class imbalance. In contrast, other tests showed minimal differences across balanced and imbalanced cases.

\textbf{Robustness to Outliers:} This criterion reflects a test’s sensitivity to extreme raters (e.g., strong over- or under-raters). Ideally, tests should not be unduly influenced by a small number of outliers. The {\it Majority raters} test was the only one that became even stricter in the presence of outliers. Other tests, {\it Average $\kappa$}, {\it Average AC1}, and {\it IRA vs. AI-Consensus AC1} and {\it IRA vs. AI-Consensus $\kappa$} tests, showed slight sensitivity to outliers (see Appendix \ref{app:app2}, Fig.~\ref{outlier}), allowing non-experts to pass more easily, though the effect was minor. The {\it Any rater} test was generally permissive and passed nearly all raters regardless of context and was therefore uninformative under the experimental conditions examined. {\it All raters} was overly strict and unaffected by outliers, but it rejected too many raters regardless. In the outlier scenario examined, {\it Pairwise MCC} and {\it AUC} tests passed all non-experts, indicating little sensitivity to extreme rater behaviour under these conditions.

\textbf{Missing Data Resilience:} Only one test is suitable when annotations are missing without any modifications: the {\it Average $\kappa$} when Krippendorff’s $\alpha$ is used in place of Fleiss’ $\kappa$; both yield equivalent results, while Krippendorff’s $\alpha$ supports missing entries. 

\subsection{Framework Application: A Published SVM-Based Detector}
\label{sec:caseStudy}
To complement the synthetic evaluations above, we applied our complete framework to a publicly available, SVM-based, neonatal seizure detection model with released code, trained and tested on the Helsinki dataset \cite{b8}. Although this model dates from 2019 and has since been surpassed by newer architectures, it remains, to our knowledge, the only neonatal seizure detection model with open code, making it a valuable real-world test case for our framework. Applying our framework yielded a leave-one-out cross validation with AUC = 0.908, accuracy = 0.942, sensitivity = 0.732, specificity = 0.967, PPV = 0.727, NPV = 0.968, MCC = 0.697, and seizure burden agreement 0.790. While the high AUC alone would suggest strong performance, the moderate MCC and sensitivity indicate a less optimistic performance, consistent with the metric behaviour described in this paper. Applying the {\it Average $\kappa$} human--expert equivalence test, the model did not satisfy the non-inferiority criterion ($\Delta\kappa \text{ }p5 = -0.176$, required margin $= -0.077$), despite a previous validation study reporting non-inferiority using a different evaluation criterion. The model also failed the {\it Majority raters}, {\it All raters} and {\it Any rater} tests, but passed the {\it IRA vs. AI-Consensus $\kappa$} test, mirroring the relative strictness of these tests observed in our experiments.

\section{Discussion}
The AUC, while widely reported in seizure detection studies \cite{b5, b6, b7, b8, b11, b12, b13, b14, b15, b16, b17}, fails to reflect true performance degradation under high class imbalance. Even in scenarios with a high FP count and a dramatic drop in PPV, the AUC remained artificially high because it only reflects sensitivity and specificity. This issue, while previously noted in the general ML literature\cite{b19, b20, b21}, is rarely discussed in the seizure detection context and reinforces the need for metrics that account for class distribution and its real-world implications. Metrics such as MCC and PCC, which incorporate all elements of the confusion matrix, offer a more robust and accurate evaluation, especially important in long-term EEG monitoring where seizures are rare, and performance overestimation may lead to misleading clinical conclusions. Still, summary metrics alone hide error types: in seizure detection, FNs (missed seizures) may be more clinically significant than FPs. Thus, reporting sensitivity, specificity, PPV, and NPV alongside summary metrics is strongly recommended, since it provides a more complete evaluation. 

Seizure burden provides a clinically interpretable, duration-based complement to sample-based performance metrics, and Fig.~\ref{auc} shows that seizure burden correlation decreases with increasing FP/TP ratio in parallel with MCC and PCC, whereas the AUC remains constant. It does not replace a balanced metric: compensating FPs and FNs can leave total burden approximately correct while sample-level agreement is poor. It should therefore be reported alongside, rather than instead of, a balanced metric.

Event-based metrics, including event-based sensitivity and FD/h, provide complementary clinical information on whether seizure episodes are detected and how often false detections occur. However, conventional event-based definitions do not fully capture event duration or seizure burden: short and long seizures contribute equally to sensitivity, and FD/h omits the cumulative duration of false detections. Models with similar event-based results may therefore differ substantially in total seizure burden. We recommend reporting event-based metrics alongside sample-based metrics and seizure burden, with the event-matching criteria and post-processing rules clearly specified. In practice we suggest a staged use: sample-based metrics and human–expert equivalence testing provide the primary technical assessment of model capability, and event-based metrics are then reported as a complementary, clinically oriented characterisation once acceptable sample-based performance has been established. This ordering is not intended to discourage event-based reporting, which remains standard in clinical and regulatory contexts, but to ensure that reported event counts are interpreted together with the sample-level performance that determines what those counts represent.
%Event-based metrics (e.g., event-based sensitivity, FD/h) are often used to align evaluation with clinical interpretations, but they fail to account for seizure burden, a key indicator of seizure severity. Two models may achieve identical event-based scores but differ significantly in total seizure time. In contrast, sample-based metrics better reflect seizure burden and provide more reliable and clinically informative performance insights.
%Given these limitations, event-based metrics are best reported only as a complementary, clinically intuitive summary - for instance to communicate expected seizure counts or durations - once sample-based metrics and seizure burden correlation already indicate satisfactory performance, rather than as a primary tool for evaluating or comparing models. This ensures that the accessibility of event-based reporting does not come at the cost of an accurate model assessment.

Our analysis of consensus formation strategies revealed a critical tradeoff between annotation confidence and data retention. Unanimous consensus yields high-confidence labels but discards a large portion of data as rater count increases, making it suitable only when agreement is strong. Regardless, it discards all disagreements while retaining the potentially easier-to-label samples. Majority consensus, while less strict, preserves all the data but may contain significant disagreement, therefore undermining consensus. These findings imply that the choice of consensus strategy should depend on the number of raters and the desired level of label reliability. Neither approach is ideal; however, unanimous consensus can be misleading by excluding all disagreement, and the proportion of discarded data should always be reported.

Despite increasing interest in ML models for seizure detection, there remains a lack of standardisation in how performance is evaluated. Many studies rely on a narrow set of summary metrics, most commonly AUC, which can obscure critical aspects of model behaviour, particularly in highly imbalanced settings such as neonatal EEG. In this context, human--expert equivalence tests offer a valuable complement to standard metrics by directly assessing whether an AI system performs at the level of a trained human rater, an important benchmark to evaluate against. These tools can help ensure that models not only perform well statistically but also meet the standards of real-world clinical interpretation. However, human--expert equivalence tests remain inconsistently applied in the literature. Given the significance of the claim of expert-level equivalence (in neonatal \cite{b9, b18} and adult EEG \cite{b22, b30}\footnote{With the same claims for neonatal EEG on their website \url{https://www.persyst.com/technology/seizure-detection/}}), it is important to establish the validity of these tests. Among these studies, only \cite{b9} used the {\it Average $\kappa$} test, which showed the strongest and most consistent performance among the tests evaluated under our experimental conditions, while others relied on the {\it Any rater} test \cite{b18}, {\it IRA vs. AI-Consensus AC1} test \cite{b22}, and also a variation of the pairwise test proposed in \cite{b23}. We also note that the pairwise test used in \cite{b23} has additional limitations when event-based sensitivity and FD/h are interpreted in isolation, because these measures do not fully characterise the duration or temporal distribution of detection errors. Moreover, the method incorrectly assumes independence between sensitivity and FD/h. To address this, such evaluations should instead consider joint performance or rely on a single balanced metric, as proposed in this work.

Among the human--expert equivalence tests evaluated, the multi-rater statistical Turing test with the {\it Average $\kappa$} variant consistently demonstrated the strongest performance and robustness across the experimental conditions examined. It achieved the highest $\mathcal{A}_W$ across all controlled dataset groups (D1--D4), demonstrating robustness to annotation bias, rater composition, and class imbalance. Although slightly sensitive to imbalance and outliers, it reliably distinguished experts from non-experts across all tested conditions. Its robustness was further enhanced by the possibility of substituting Fleiss’ $\kappa$ with Krippendorff’s $\alpha$ to support missing data. Compared to stricter variants (e.g., {\it All raters}) or lenient ones (e.g., {\it Any rater}), it avoids false rejection of experts while preventing overestimation of AI capabilities in expert-level tasks. These findings were further supported by our case study applying the full framework to a published SVM-based model (Section \ref{sec:caseStudy}): the model's moderate MCC and sensitivity pointed to substantially weaker performance than the AUC alone would suggest, and it failed the {\it Average $\kappa$} equivalence test despite an earlier non-inferiority claim based on a different criterion. This real-world example reinforces that both metric choice and equivalence-test choice can significantly change conclusions about a model's readiness for clinical use. This does not imply that the alternative procedures are invalid in general; rather, they exhibited different strictness and sensitivity characteristics under the experimental conditions examined, and their behaviour in other rater populations and datasets remains to be established.%\textcolor{blue}{We note that Method A models rater and AI bias as an aggregate, spatially uniform tendency rather than explicitly modelling content-driven disagreement arising from morphological EEG patterns that may resemble seizure activity. However, the multi-rater statistical Turing test relies on the relative change in agreement ($\Delta\kappa$) when an AI substitutes for a human rater, rather than on absolute agreement levels. Since both simulated raters and the simulated AI are generated under the same modelling assumptions in our framework, this relative comparison is expected to remain unaffected by the simplification, even though a more content-driven error model, where both human raters and an AI trained on their annotations would plausibly share confusion around the same morphologically ambiguous segments, might shift absolute agreement levels. Explicitly modelling such correlated disagreement remains an important direction for future work.}

A notable example was the sensitivity of the {\it Average AC1} test to class imbalance. Although Gwet’s AC1 is designed to avoid the paradox of low kappa for high agreement due to imbalance, it became progressively less responsive to minority-class disagreement in the controlled experiments examined here. At majority-to-minority ratios above approximately 10:1, AC1 remained above 0.9 even when no minority-class sample was correctly classified (Appendix \ref{app:app1}). This behaviour propagated into the AC1-based procedures examined here, as reflected in the reduced performance of {\it Average AC1} in D2 and D4. Therefore, the use and interpretation of AC1 in rare-event annotation tasks should account for its potential insensitivity to minority-class errors under severe class imbalance.

This study combines controlled synthetic experiments with real-data analyses, and its conclusions should be interpreted accordingly. The consensus-strategy analysis (Section~\ref{sec:sec32}), the synthetic datasets comparison with the Helsinki and Cork annotations (Fig.~\ref{synthetic_annotations_compared_to_real}), and application to a published neonatal seizure detector (Section~\ref{sec:caseStudy}) provide real-data grounding. In contrast, the metric-behaviour and expert-equivalence experiments (Sections~\ref{sec:sec31} and \ref{sec:sec33}) rely on synthetic annotations because they require independent control of class imbalance, rater number, ground truth, and rater composition beyond what the available fixed datasets permit. These experiments operate on per-sample binary annotations: sample-based metrics are determined by the resulting TP, TN, FP, and FN counts, while agreement tests depend on the alignment of binary labels across raters. 
%Consequently, reproducing the underlying EEG morphology or identifying whether each disagreement arose from an artefact, morphological ambiguity, boundary uncertainty, or another interpretive mechanism is not required for the specific annotation-level procedures evaluated here. Method A instead represents their observable consequences through systematic over- and under-annotation and between-rater variability. 
%Nevertheless, Fig.~\ref{synthetic_annotations_compared_to_real} demonstrates representational capacity rather than independent external validation because the same real annotations informed the calibration, and transfer to annotation structures not represented by the generator remains to be established. 
Event-based metrics were intentionally excluded because, unlike the sample-based procedures examined here, they depend on realistic event durations, temporal clustering, and event boundaries, which the framework does not generate. Finally, the real-model analysis was limited to the only identified neonatal detector with publicly available code and compatible trained models for the multi-rater Helsinki dataset. Broader validation therefore depends on the release of further compatible multi-rater datasets, trained models, private evaluation, or model predictions.

The primary goal of ML models for neonatal seizure detection is to support clinical decision-making. Fair and rigorous evaluation is therefore not optional; it is a prerequisite for responsible translation to bedside use. Neglecting to consider appropriate evaluation metrics or annotation variability can lead to the adoption of models that do not generalise, misinform clinical workflows, and ultimately erode trust in the potential of AI tools. To prevent this, based on our analysis, we propose the following reporting recommendations for evaluating seizure detection algorithms: 
\begin{enumerate}
    \item At least one balanced metric (e.g., MCC or PCC),
    \item Sensitivity, specificity, PPV, and NPV to clarify error types,
    \item Seizure burden agreement, including correlation between model-estimated and reference burden across hourly windows (assessed using PCC), together with the corresponding burden estimates expressed in minutes per hour,
    \item \textit{Average $\kappa$} human--expert equivalence test,
    \item Reporting all metrics on held-out validation sets.
\end{enumerate}
These practices promote fairer model comparison, reproducibility, and real-world applicability. While the present study focuses on methodological evaluation rather than clinical suitability, deployment, or clinical monitoring, the choice of evaluation metrics has important practical implications. Metrics that are relatively insensitive to FPs may underestimate the burden of unnecessary alarms in NICU monitoring, whereas balanced metrics, reported together with sensitivity, specificity, PPV and NPV, provide a more informative characterisation of model behaviour. However, acceptable false-alarm rates, clinical decision thresholds, and workflow requirements depend on the intended clinical use and cannot be determined from evaluation metrics alone. Establishing these requirements represents a subsequent stage of evaluation requiring prospective investigation in the intended clinical setting and continuous post-deployment surveillance. The standardised evaluation proposed here is an important prerequisite for such studies of the clinical utility and integration of AI-assisted seizure detection systems.

Although developed for neonatal seizure detection, the proposed evaluation framework is in principle transferable to other EEG and time-series detection problems characterised by annotation uncertainty, severe class imbalance, and multiple expert raters, although no non-neonatal data were analysed in this study. Similar concerns have also been raised in the adult EEG literature \cite{b33, b34}. Because the framework operates directly on annotations rather than on the underlying EEG signal, extending it to other domains primarily requires adapting the annotation characteristics, such as baseline IRA and class distribution. Application to adult EEG would require, at minimum, re-parameterisation to reflect the relevant class distribution and inter-rater agreement, and may require further adaptation to domain-specific annotation structures. This expectation follows from the structure of the framework rather than from empirical evidence, and validation on adult multi-rater EEG datasets would be required. Despite the comprehensive quantitative analyses, evaluating seizure detection models remains inherently challenging. The next stage of evaluation of clinical utility requires prospective clinical studies and ongoing post-market surveillance in diverse, operational environments to assess how AI models can positively impact patient care and workflow. 

%%%%%%%%%%%%%%%%%%%%%%%%%%%%%
%\clearpage
\appendix
\renewcommand{\thefigure}{A\arabic{figure}}
\renewcommand{\theequation}{A\arabic{equation}}
\setcounter{figure}{0}
\setcounter{equation}{0}

\section{Inter-rater agreement measures}
\label{app:app1}

\begin{figure}[!bt]
\centerline{\includegraphics[width=1.0\columnwidth]{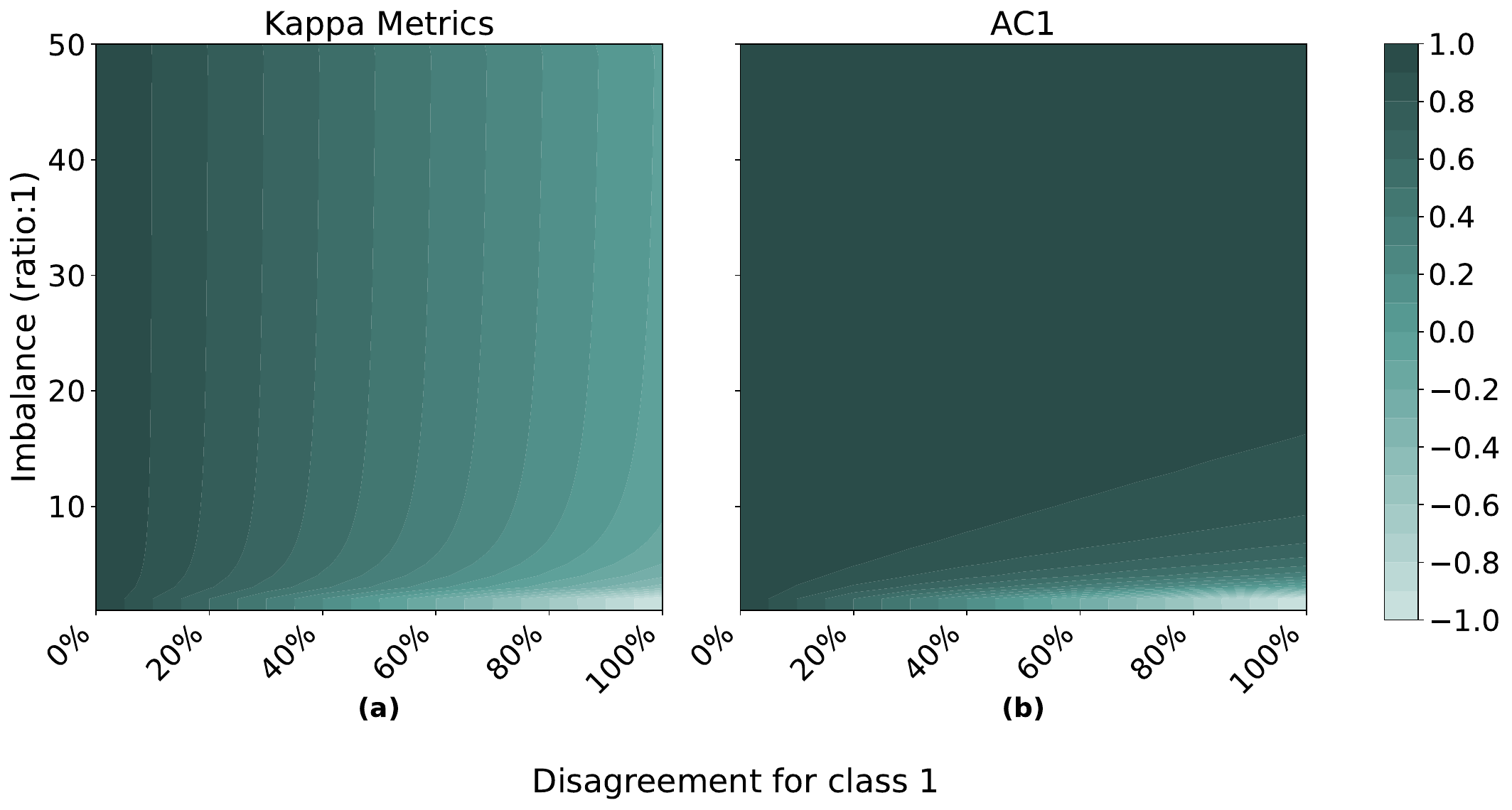}}
\caption{Effect of class imbalance and increasing disagreement on IRA metrics. Annotations generated with Method B, Variation 2. The x-axis represents the percentage of TPs (seizure samples) misclassified as FNs, with an equal number of FPs introduced to preserve the original class distribution. Cohen’s $\kappa$, Fleiss’ $\kappa$, and Krippendorff’s $\alpha$ yield identical values in this experimental setting and are shown together as “Kappa metrics" (a). Unlike these metrics, AC1 (b) becomes insensitive to disagreement under severe class imbalance ($>10:1$), producing artificially high values even when minority-class errors dominate.} 
\label{ira}
\vspace{-1em} % reduce vertical space after caption
\end{figure}

\setcounter{figure}{0}
\setcounter{equation}{0}
\renewcommand{\thefigure}{B\arabic{figure}}
\begin{figure}[!bt]
\centerline{\includegraphics[width=1.0\columnwidth]{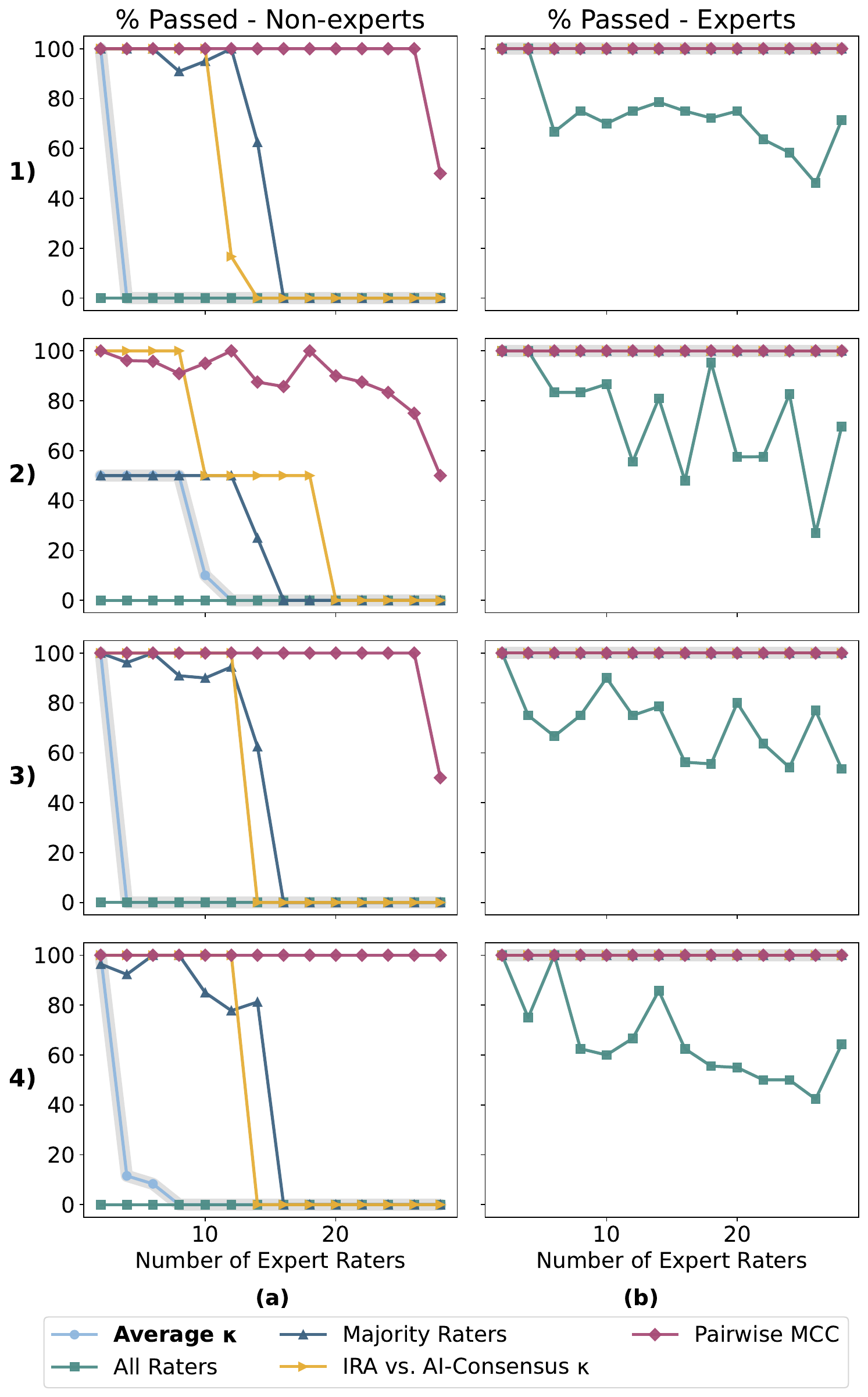}}
\caption{Performance of the evaluated human--expert equivalence tests on: 1) D1 dataset group (balanced class distribution, 1:1, with a combination of under- and over--raters), 2) D2 dataset group (imbalanced class distribution, 25:1, with a combination of under- and over-raters), 3) D3 dataset group (balanced class distribution, 1:1, with directionless non-experts), and 4) D4 dataset group (imbalanced class distribution, 25:1, with directionless non-experts). The number of expert raters increases from 1 to 29 (x-axis) while the total number of raters remains fixed at 30. The column (a) represents percentage of non-experts incorrectly passing each test, whereas the column (b) shows the percentage of experts correctly passing each test. A robust method should reject an increasing proportion of non-experts while consistently allowing experts to pass as the number of experts increases. Annotations were generated using Method A.
}
\label{D1}
\vspace{-1em} % reduce vertical space after caption
\end{figure}

\begin{figure}[!bt]
\centerline{\includegraphics[width=1.0\columnwidth]{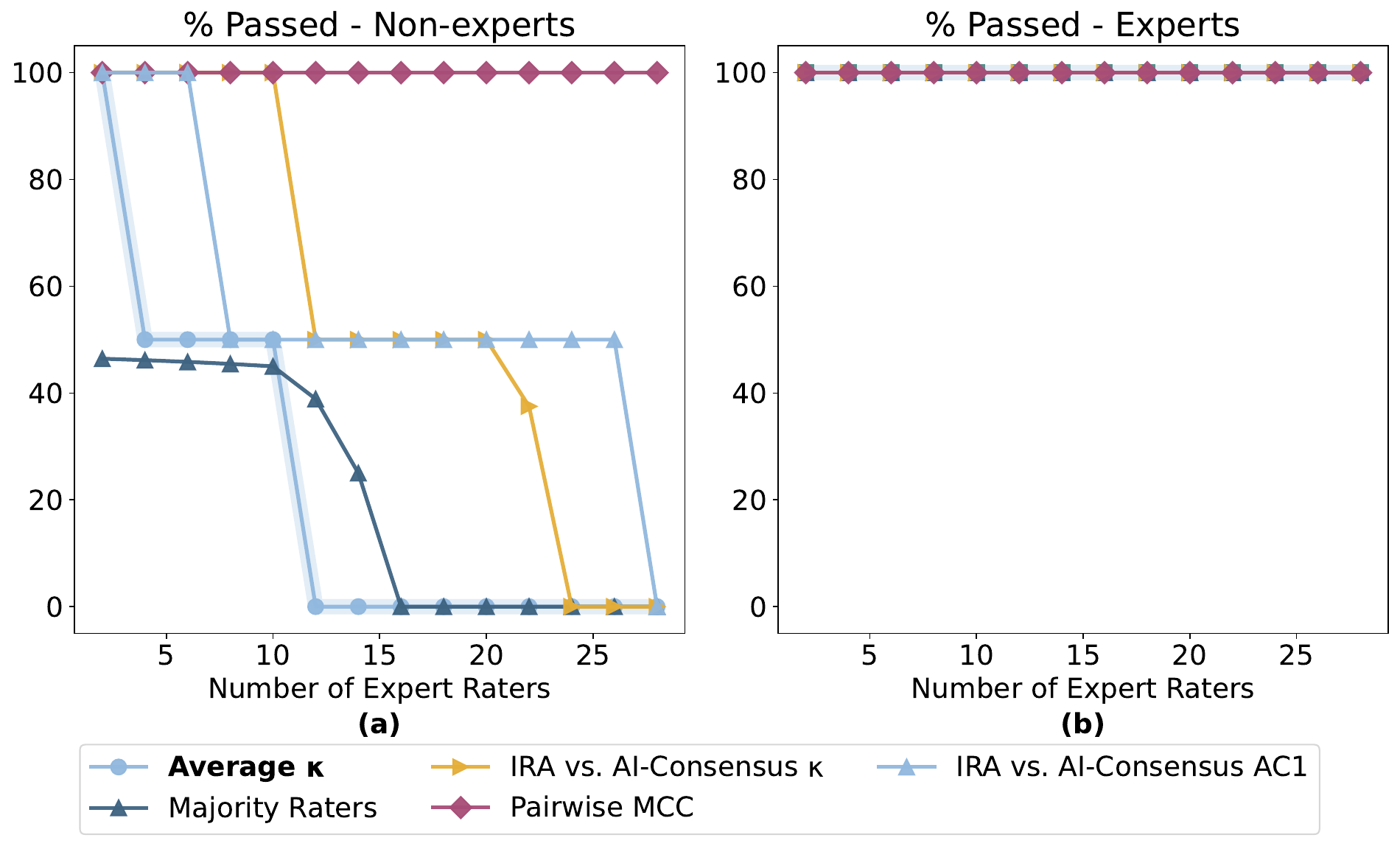}}
\caption{Effect of an outlier rater on human--expert equivalence tests. Performance of the evaluated methods on dataset group D2 after introducing a single extreme rater (strong over- or under-rater). (a) Percentage of non-experts passing each test. (b) Percentage of experts passing each test. Comparison with Fig.~\ref{D1} row 2) illustrates each method's robustness to outlier annotations.}
\label{outlier}
\vspace{-1em} % reduce vertical space after caption
\end{figure}

Consensus-based assessment disregards IRA, which is a crucial context for assessing detection models. Most agreement coefficients follow the general form:
\begin{equation}
   \kappa = \frac{P_o-P_e}{1-P_e} \label{kappa}
\end{equation}
where $P_o$ represents the observed agreement and $P_e$ represents expected agreement by chance. Different measures vary in how $P_e$ is estimated and whether they support multiple raters, missing data, or different data types.

Cohen’s $\kappa$ \cite{a1} is one of the most widely used IRA metrics, designed for comparing two raters. Fleiss’ $\kappa$ \cite{a2} generalises this to more than two raters. Both can be sensitive to class imbalance effects \cite{a3, a4}, often underestimating agreement. Gwet’s AC1 \cite{a5} attempts to correct this with a modified calculation of expected agreement, offering more stable estimates under class imbalance. For extremely imbalanced datasets, however, it has the opposite problem of overestimating agreement. While all of these expect complete, nominal data, Krippendorff’s $\alpha$ \cite{a6}, in contrast, is a more flexible extension of Fleiss’ $\kappa$ that accommodates missing data and supports a range of data types (nominal, ordinal, interval).

We used Method B and its Variation 2 to generate synthetic annotations with varying levels of disagreement among raters. The error rate ranged from $0\%$ to $100\%$, directly controlling rater sensitivity. This process was applied across datasets with increasing class imbalance (from 1:1 to 50:1), enabling a systematic evaluation of how IRA metrics respond to both rater disagreement and class distribution.

Fig.~\ref{ira} shows that without imbalance, all metrics responded approximately linearly to the percentage of disagreement and spanned the full range from $-1$ to $1$. As class imbalance increases, this linear relation remains; however, we observe a narrowing of the range from $-1$ to $1$, especially for metrics favouring the majority class. While “Kappa metrics” (Cohen’s $\kappa$, Fleiss’ $\kappa$, and Krippendorff’s $\alpha$ are all equivalent in this setup) retain a large range from approximately 0 to 1 at a class imbalance of 50:1, the AC1 range completely collapses once the class imbalance exceeds 10:1 and becomes independent of $\%$ disagreement. Specifically, even in scenarios where no samples from the minority class were correctly classified, the AC1 score remained above $90\%$, while the Kappa metrics exhibited a noticeable decline in performance, and their values reached zero, indicating that one class is entirely misclassified. This is a consequence of AC1's tendency to favour the majority class, leading to overly optimistic outcomes. Under the severely imbalanced configurations examined here, these results indicate that AC1 may be unsuitable as the sole IRA measure because it insufficiently reflects minority-class disagreement.

\section{Results of Human--Expert Equivalence Tests}
\label{app:app2}
To analyse the performance of human--expert equivalence tests, the evaluation was framed as a binary classification task, where the goal was to distinguish expert raters (positive class) from non-expert raters (negative class) based on pass/fail outcomes. Fig.~\ref{D1} illustrates the performance of selected tests across four dataset groups (D1–-D4), as the number of expert raters increases from 1 to 29 (x-axis), with the total number of raters fixed at 30. Each row shows the percentage of non-experts and experts who passed the test at each expert proportion. This setup enabled a qualitative comparison of test behaviour, including sensitivity to class imbalance, by comparing D1 with D2 and D3 with D4. As non-experts dominate when expert counts are low, it is expected that some non-experts may pass early on. However, robust tests should increasingly reject non-experts as the proportion of experts rises.

Figure \ref{outlier} presents the performance of selected tests when a single extreme rater (e.g., a strong over- or under-rater) is introduced into dataset group D2. A comparison between Fig.~\ref{D1} row 2) and Fig.~\ref{outlier} reveals that the presence of such an outlier slightly increases the likelihood of non-experts passing the tests, indicating a mild sensitivity to extreme ratings.

%\vspace*{-1.5em}
%\clearpage

\end{document}